\documentclass[journal]{IEEEtran}

\usepackage{amsmath}
\usepackage{amsfonts}
\usepackage{algorithm}
\usepackage[noend]{algpseudocode}
\usepackage{algorithm}
\usepackage{booktabs}
\usepackage{array}
\usepackage[numbers]{natbib}
\usepackage{multirow}
\usepackage[labelformat=simple]{subcaption}
\usepackage{graphicx}
\usepackage{listings}
\usepackage[utf8]{inputenc}
\usepackage[T1]{fontenc}
\usepackage{fancyvrb}

\DeclareMathOperator{\mean}{mean}

\begin{document}
\bstctlcite{IEEEexample:BSTcontrol}

\title{Playtesting: What is Beyond Personas}

\author{
    \IEEEauthorblockN{Sinan~Ariyurek\IEEEauthorrefmark{0},~Elif~Surer\IEEEauthorrefmark{0},~Aysu~Betin-Can\IEEEauthorrefmark{0}}
    \IEEEauthorblockA{\IEEEauthorrefmark{0}
    \\Graduate School of Informatics
    \\Middle East Technical University
    \\06800, Ankara, Turkey
    \\\{sinan.ariyurek, elifs, betincan\}@metu.edu.tr}
}


\maketitle


\begin{abstract}
Playtesting is an essential step in the game design process. Game designers use the feedback from playtests to refine their designs. Game designers may employ \textit{procedural personas} to automate the playtesting process. In this paper, we present two approaches to improve automated playtesting. First, we propose \textit{developing persona}, which allows a persona to progress to different goals. In contrast, the procedural persona is fixed to a single goal. Second, a human playtester knows which paths she has tested before, and during the consequent tests, she may test different paths. However, Reinforcement Learning (RL) agents disregard these previous paths. We propose a novel methodology that we refer to as Alternative Path Finder (APF). We train APF with previous paths and employ APF during the training of an RL agent. APF modulates the reward structure of the environment while preserving the agent's goal. When evaluated, the agent generates a different trajectory that achieves the same goal. We use the General Video Game Artificial Intelligence (GVG-AI) and VizDoom frameworks to test our proposed methodologies. We use Proximal Policy Optimization (PPO) RL agent during experiments. First, we compare the playtest data generated by developing and procedural persona. Our experiments show that developing persona provides better insight into the game and how different players would play. Second, we present the alternative paths found using APF and argue why traditional RL agents cannot learn those paths.
\end{abstract}


\IEEEpeerreviewmaketitle

\section{Introduction}\label{sec:introduction}

Game designers envision how a game will work during a play through. As the game develops, it becomes increasingly difficult to predict how players will interact with the game. Playtesters help out this process by providing feedback by playing the game. However, human playtesting introduces latency and additional costs to the process. Therefore, researchers proposed methods to automate the playtesting process~\cite{Powley:2016}~\cite{Gudmundsson:2018}~\cite{Roohi:2020}. 

Additionally, the playtesting process may employ players with distinct playstyles. These players will respond to the game differently, and they will generate different play traces. The game designer can use these play traces to shape her game. In order to automate playtesting with different players, researchers replaced these playtesters with procedural personas. A procedural persona describes an archetypal player's behavior. Researchers used personas to playtest a Role-Playing Game~\cite{Holmgard:2018} and a Match-3~\cite{Mugrai:2019} game. As a result, personas enabled distinct playstyles and helped to playtest a game like distinct players. 

In order to realize the personas using RL agents, researchers used a utility function~\cite{Holmgard:2014:q} to define the decision model of a persona. This utility function was used as the reward function of the Q-Learning agents. However, this replacement makes the agents bound to the utility function. Since the utility function is tailored for a specific decision model, the behavior of these agents is constant throughout the game. Therefore, the procedural personas approach is not flexible enough to create personas with developing decision models. For example, a player may change her objectives while playing the game. Consequently, the decision model of this player cannot be captured by a utility function.

Bartle~\cite{Bartle:2005} presents examples of these changes that a player can undergo while playing a Massively Multiplayer Online Role-Playing Game. We believe that the change in the playstyle occurs after accomplishing a goal. For example, a player may start a game by opening the treasures to find a required item and then killing monsters. This player chooses her actions like a Treasure Collector until she finds the desired item and becomes a Monster Killer. We propose a sequence of goals to model the decision-making mechanism of this player. The sequence-based approach was previously used in automated video game testing agents~\cite{Ariyurek:2019} and was found more practical than non-sequence-based approaches.

The developing persona model consists of multiple goals that are linked. Each goal consists of criteria and a utility function. The utility function serves the same purpose as in procedural personas. The criteria determine until which condition the current goal is active. When the current goal criteria are fulfilled, the next goal becomes active. The agent plays until the last goal criterion is fulfilled or until the end of the game. The game designer sets the criteria and utility functions of each goal. The goal structure enables the creation of dynamic personas. Additionally, this approach gives a more granularized control over a persona. The game designer can create variations of Monster Killer by setting different criteria. In order to playtest a casual Monster Killer, the game designer may set a health threshold as the criterion; and to playtest a hardcore Monster Killer, the game designer may set the percentage of monsters killed as the criterion.

Furthermore, the game designer may envision a game with various endings. In order to playtest her game, she utilizes an agent that behaves like an Exit persona and exercises this agent multiple times. Then, the game designer analyzes the trajectories generated by this agent and sees that all the trajectories provide data for only one of the possible endings. On the other hand, a human playtester would have generated trajectories that cover various endings. Thus, the shortcoming of automated playtests is not caused by the Exit persona but by the inherent nature of RL algorithms. RL algorithms such as Deep Q-Network (DQN)~\cite{Mnih:2015}, Proximal Policy Optimization (PPO)~\cite{Schulman:2017}, and Monte Carlo Tree Search (MCTS)~\cite{Browne:2012} disregard the previous trajectories. Consequently, even if we train an agent with any of these algorithms and then evaluate the agent, and repeat this process numerous times, all the generated trajectories would be similar most of the time. However, the trajectories may be different due to the following reasons a) the random initialization of the Neural Network (DQN and PPO) b) $\epsilon$-greedy policy of DQN c) stochasticity of MCTS d) the game's nondeterminism. The critical point is that even if the agent generates a distinct trajectory, this result is not by design but by random chance.

Exploration methods in RL improve the agent's policy by motivating the agent to explore the environment. As the agent explores an environment, the agent improves its policy. The researchers proposed methods to motivate the agent to explore less visited states~\cite{Bellemare:2016}~\cite{Fu:2017}~\cite{Pathak:2017}. Compared to the traditional exploration methods such as $\epsilon$-greedy, where exploration is achieved through randomness, these modern algorithms entice exploration logically. These algorithms learn to distinguish the unvisited states from the visited states, consequently, these algorithms guide the agents to less-visited states. As a result, exploration methods vastly improved the agent's score, such as Montezuma's Revenge~\cite{Bellemare:2016}.

On the other hand, APF knows the previous trajectories and guides the agent to learn to play differently from previous ones. For this purpose, APF penalizes the agent when the agent visits a similar state and rewards the agent when the agent visits a different state, compared to the states in the previous trajectories. APF employs the state comparison algorithms used in exploration algorithms. These state comparison algorithms are the backbone of exploration research, and researchers tested these algorithms in multiple games. We show how we build the APF framework to generate new and unique playtests and how APF augments any RL agent.

In this paper, we list the contributions as follows. Our first contribution is the developing persona. The developing persona is more flexible and capable than the current persona models. We show how game designers can utilize the developing persona to empower the playtesting. Our second contribution is the Alternative Path Finder. We present a generic APF framework that can augment every RL agent. We use the GVG-AI~\cite{GVGAI:2019} and VizDoom~\cite{VizDoom:2016} environments to demonstrate our proposed methodologies.

This paper is structured as follows: Section \ref{sec:related_research} describes the examples and methodologies of related research. We grouped the related research into four subsections: Playtesting, Personas in Playtesting, Automated Playtesting, and Exploration Methods in Reinforcement Learning. The developing persona is based on the first three subsections. Next, APF is founded on the Exploration Methods in Reinforcement Learning. Our proposed methodology that consists of developing persona and APF is presented in Section \ref{sec:methodology}. Section \ref{sec:experiments} describes our experimentation setup and Section \ref{sec:results} presents the results of these experiments. Section \ref{sec:discussion} discusses the outcomes of the strategies used, their contributions and limitations. Lastly, Section \ref{sec:conclusion} concludes this paper.

\section{Related Research}\label{sec:related_research}

\subsection{Playtesting}
Playtesting is a methodology used in the game design process. Playtesters test a game, and feedback is collected from these playtesters. The game designers use this feedback to improve their game. As this process requires a human effort, researchers proposed methods to automate game playtesting. Powley et al.~\cite{Powley:2016} coupled automated playtesting with a game development application. Gudmundsson et al.~\cite{Gudmundsson:2018} trained a convolutional neural network to predict the most humane action in Candy Crush, and they used this network to assess level difficulty. Roohi et al. ~\cite{Roohi:2020} used RL and a population model to determine level difficulty for Angry Birds Dream Blast. These approaches derive the automated playtesters from an individual player archetype. Nevertheless, during a playtest, there can be various playtesters resembling a different player archetype.

\subsection{Personas in Playtesting}
In playtesting, personas provide game designers information about how different player archetypes would play the game. Persona is a fictional character that represents a user type. Bartle~\cite{Bartle:1996} introduces a taxonomy of personas that are identified from a Multi-user Dungeon Game. The author acknowledges these four distinct personas as Socializers, Explorers, Achievers, and Killers. The author introduces a graph with axes that maps the players' interest in a persona. Bartle~\cite{Bartle:2005} extends this research by introducing development sequences for personas. The development sequences reveal how and why a player may change to a different persona. Tychsen and Canossa~\cite{Tychsen:2008} present a study on collecting game metrics and how different personas can be identified by these metrics. The authors present the personas of the game Hitman Blood Assassin. The game identifies these personas: \textit{Mass Murderer}, \textit{Silent Assassin}, \textit{Mad Butcher}, and \textit{The Cleaner}. They argue that a persona can be recognized using the metrics collected from a play trace. These approaches focus on identifying different personas in a game.

\subsection{Automated Playtesting}
In order to automate the playtesting, researchers proposed techniques to realize the decision model of a persona. Holmg{\aa}rd et al.~\cite{Holmgard:2014:q} used a utility function to realize the decision model of a persona. This utility function is used as the reward function for the Q-Learning agent. The agents are exercised in an environment called MiniDungeons. The agents produced play traces as if they are of a specific persona. Holmg{\aa}rd et al.~\cite{Holmgard:2014:e} extended their previous work by substituting the Q-Learning agents with a neural network. The inputs to the neural network were hard-coded, handpicked parameters. The authors used a genetic algorithm to find the weights of this neural network. They called their new method `evolved agent'. Evolved agent required less training than the Q-Learning agent and was able to generalize to other levels better. Holmg{\aa}rd et al.~\cite{Holmgard:2015} upgraded the environment to MiniDungeons 2. In this study, the authors proposed to generate personas using MCTS agents that use their proposed utility function. Their reasoning for using MCTS, especially Vanilla MCTS~\cite{Browne:2012}, was to provide faster data to the game designer. In Q-Learning and Evolved agents, these agents have to be trained first. Holmg{\aa}rd et al.~\cite{Holmgard:2018} extend the MCTS by improving the selection method of MCTS\@. In their previous study, the authors state that the Mini Dungeons 2 game was too complex for Vanilla MCTS\@. Therefore, they model a new selection phase that is specifically tailored towards a specific persona. They accomplish this by evolving the UCB formula by a genetic algorithm. The authors crafted the fitness function of each persona. This fitness function also determined the fitness function of the evolutionary algorithm. The evolved UCB formula improved their results among every persona. Silva et al.~\cite{Silva:2017} used personas to playtest the Ticket to Ride board game. The authors designed four different competitive personas to play the board game. The authors handcrafted a set of heuristics for each persona. They showed that personas revealed useful information that the game rules did not provide rules for two situations. Mugrai et al.~\cite{Mugrai:2019} employed four different personas for Match-3 games. These personas are \textit{Max Score}, \textit{Min Score}, \textit{Max Moves}, and \textit{Min Moves}. The authors showed that these four personas could give the game designer valuable information about a level.

The main drawback of persona research is the utility function. First, the utility function is static and stays constant throughout the game. Therefore, the game designers cannot model players with development sequences~\cite{Bartle:2005}. Second, depending on the level layout, personas can execute a similar sequence~\cite{Holmgard:2018}. Hence, the synthetic playtesters would provide ineffective feedback. Lastly, synthetic playtesters are realized using RL agents. Since RL agents optimize the total accumulated reward, synthetic playtesters would not test all playable paths.

\subsection{Exploration Methods in Reinforcement Learning}
An RL agent explores the environment to learn which action yields the highest reward in a state. In order to learn this policy, the RL agent has to explore the environment. Intrinsically motivating an RL agent to explore novel states is an exploration problem. The researchers proposed different ways to make agents explore distinct states of the environment. Count-based approaches reward the less-visited states more than frequently visited states. Therefore, the agent becomes inclined to visit the less visited states. The count is formulated using a density model~\cite{Bellemare:2016}, a neural density model~\cite{Ostrovski:2017}, a hash table~\cite{Tang:2016}, and exemplar models~\cite{Fu:2017}. Another proposed approach is to augment the reward function by measuring the agent's uncertainty about the environment. Researchers measured the uncertainty using bootstrapped DQN~\cite{Osband:2016}, state-space features~\cite{Pathak:2017}, and error of a neural function~\cite{Burda:2018}. Additionally, researchers proposed approaches that explore the state space by optimizing the state marginal distribution to match a target distribution~\cite{Lee:2019}. These exploration proposals intelligently incite the agent to explore the environment. The goal of exploration is not to find a unique way of playing but to find the best path every time we execute the RL agent. However, these methods can differentiate between similar states and new states. We base our APF proposal based on this accomplishment.

\section{Methodology}\label{sec:methodology}

In this paper, we address the shortcomings of the procedural persona with a multi-goal oriented persona, the developing persona. Additionally, we recognize there may be alternative playtests that a persona may produce. We propose APF to discover those playtests.

In the following subsections, first, we introduce the developing persona. Afterward, we present the necessity for an APF and introduce the foundation of APF. Next, we show how we use the techniques in exploration field to implement the APF. Finally, we describe how to use APF with an RL agent.

\subsection{Developing Persona}\label{m:gbp}

A persona reflects an archetypal player's decision model. In order to realize a persona, first, the persona's decision model should be translated to game conditions. Second, an actor should play according to this translation. Researchers~\cite{Holmgard:2018}~\cite{Mugrai:2019} proposed using a utility function to map the decision model to game conditions. This utility function replaces the reward mechanism of the environment and provides a tailored reward mechanism for each persona. Researchers~\cite{Holmgard:2018}~\cite{Mugrai:2019} used RL agents as actors. Consequently, these RL agents are akin to synthetic playtesters that represent the decision model of a persona. These playtesters, procedural personas, represented various personas such as the Monster Killer, Treasure Collector, and Exit personas. In this paper, we extend the procedural persona framework by introducing a multi-goal persona.

We propose a multi-goal persona to generate a more customizable playtester. We have two reasons that a multi-goal persona would be beneficial for game designers. First, the game designer does not have granular control over the personas. For example, the game designer may want to playtest a monster killer persona that kills monsters until its health drops below a certain percent. However, when to cease killing monsters was left to the RL agent, and the game designer had little control over these decisions \cite{Brown:2015}. Second, the previous approaches do not allow development in persona. Though procedural personas may realize the persona archetypes that Bartle~\cite{Bartle:2005} presented, procedural personas cannot realize the development sequences that Bartle also presents. For example, if the goal of the procedural persona is killing monsters, the procedural persona will always be a Monster Killer.

A multi-goal persona is a procedural persona with a linked sequence of goals rather than a single utility function. A goal contains a utility function and a transition to the next goal. If there is a single goal in the sequence, there is no need to define the transition. Hence, a goal-based persona with a single goal is equivalent to a procedural persona. The transition connects the goals, and the transition occurs depending on the criteria. Game designers determine the criteria, and criteria hold conditions related to the game. For example, a criterion can be killing 50\% of the monsters or exploring 90\% of the game or having health less than 20\% or the combination of these conditions. The developing persona maintains knowledge of interactions such as how many \textit{Monsters}, \textit{Treasures} have been killed or collected. Next to the interactions, the developing persona knows how much of its health is left. Developing persona uses this knowledge to check whether the current criteria are fulfilled. When all of the criteria of the current goal are fulfilled, the next goal becomes active. When there are no more goals, the training or the evaluation of the goal-based persona ends.

In this section, we have described the ``sudden'' transitions between goals, the previous goal becomes inactive, and the next goal becomes active immediately. However, this transition could also be ``fuzzy''. The current goal and the next goal can be active simultaneously. A possible implementation of fuzzy transition may use the criteria fulfillment percentage. For example, when the criteria are completed at least 50\%, the next goal could become active while not deactivating the current goal. The persona would be rewarded from both of the utility functions. Whenever the persona fulfills the current goal completely, the next goal becomes the only active goal. Consequently, a fuzzy transition would create a smoother progression of playstyles.

\begin{figure}[]
    \centering
    \includegraphics[width=0.8\columnwidth]{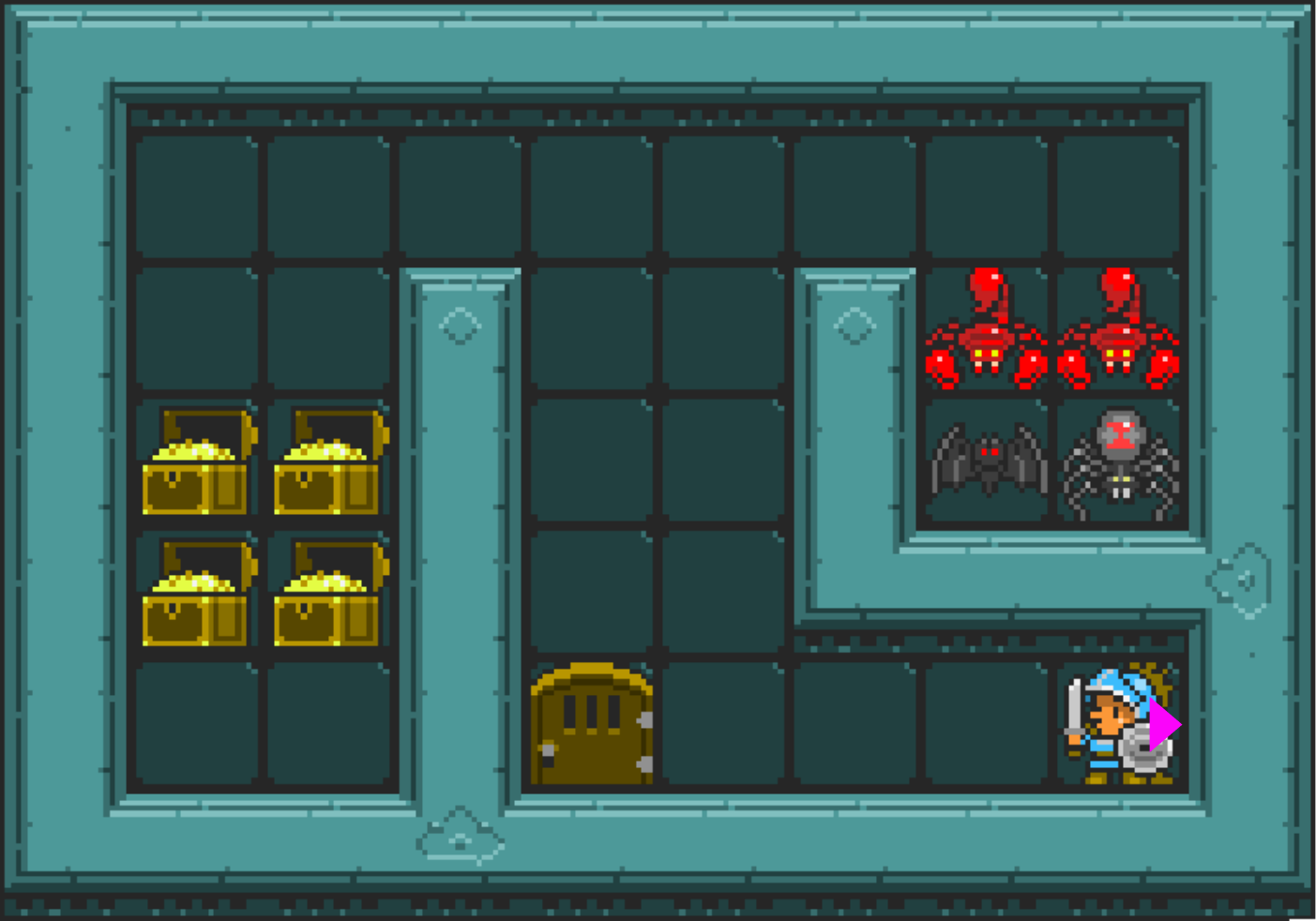}
    \caption{An example level created by GVG-AI framework.}
    \label{fig:example_game}
\end{figure}

In \figurename{ \ref{fig:example_game}}, we created an example level to demonstrate the goal-based personas. In this example, the \textit{Avatar} situated at bottom right corner can execute the following actions \textit{Pass}, \textit{Attack}, \textit{Left}, \textit{Right}, \textit{Up}, and \textit{Down}. The direction of the \textit{Avatar} is shown by a pink triangle. If the direction of the \textit{Avatar} and the action align, the \textit{Avatar} moves one space in that direction, else the \textit{Avatar} changes direction. When \textit{Avatar} executes \textit{Attack}, the \textit{Avatar} slashes towards its direction. The \textit{Avatar} can slay Monsters by Attacking them. The monsters move randomly and kill the \textit{Avatar} if they collide with the \textit{Avatar}. There are also Treasure chests that \textit{Avatar} can pick up by simply moving over them. Lastly, when the \textit{Avatar} exits through the \textit{Door}, the game terminates successfully.

\renewcommand{\arraystretch}{1.2}
\begin{table}[h]
\centering
\captionsetup{justification=centering}
\caption{Utility weights for the goals}
\begin{tabular}{cccc}
&  \multicolumn{3}{c}{ \textbf{Goal Names}}\\ 
\cline{2-4} \\
\textbf{Game Event}    & Killer & Collector & Exit  \\
\hline
Death                  &  -1.0  &  -1.0     &  -1.0 \\
Exit Door              &        &           &   1.0 \\
Monster Killed         &   1.0  &           &       \\
Treasure Collected     &        &  1.0      &       \\
\end{tabular}
\label{table:utility_functions}
\end{table}

\begin{figure}[]
    \centering
    \includegraphics[width=0.8\columnwidth]{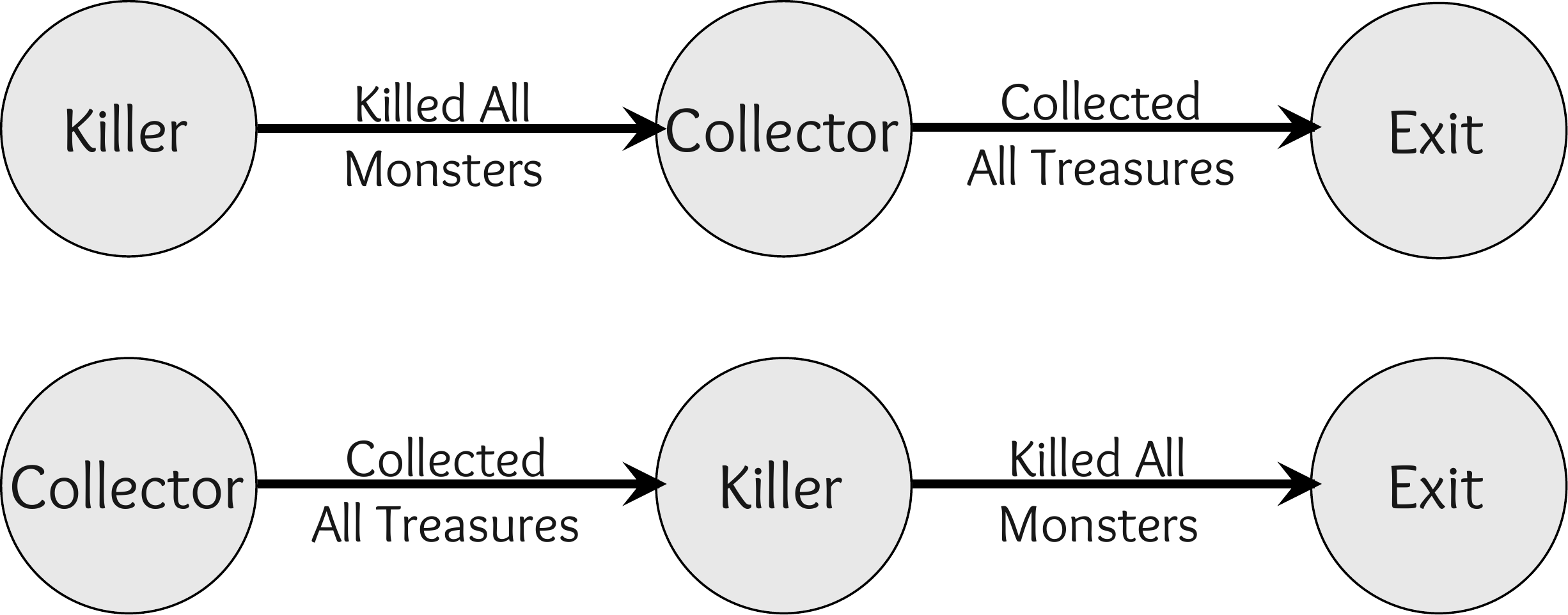}
    \caption{Developing Persona}
    \label{fig:dev_persona}
\end{figure}

A game designer may playtest a Monster Killer persona in the game shown in \figurename{ \ref{fig:example_game}} and generate the following two developing personas. First one kills the \textit{Monsters} and then collects the \textit{Treasure} as trophy. Second one collects the \textit{Treasure} hoping to gain an advantage against the \textit{Monsters} and then kills the \textit{Monsters}. In order to realize the aforementioned personas, the game designer designs two developing personas, as seen in \figurename{ \ref{fig:dev_persona}}. Next, she designs the utility functions of the goals, as seen in Table \ref{table:utility_functions}. In order to realize these personas as playtesters in a game, the game designer can employ any RL agent. When the agent finishes training, the game designer can use the agent for playtesting. The importance of developing personas is that developing personas introduces a framework to formalize how players change their goals over the course of playing a game.

\subsection{Alternative Path Finder}\label{sec:m:apf}

The actions of an RL agent are motivated based on the feedback received from an environment. As the agent is trained in an environment, the feedback will shape the agent's policy. When the training is over, the agent will behave according to the learned policy. Additionally, if we train the same agent in the same environment multiple times, the learned policies will be similar. At the end of each training, we can evaluate the trained agent in the same environment to obtain trajectories. These trajectories will be similar as the learned policies were similar. On the other hand, the game designer might be interested in seeing different playstyles.

In order to diversify the learned policies, one has to change the feedback mechanism of the environment. Procedural personas~\cite{Holmgard:2018}~\cite{Mugrai:2019} accomplish this by rewiring the feedback mechanism by a utility function. An agent representing a persona will learn a different policy than another agent that represents a different persona. However, when the game designer wants to see different playstyles within the same persona, the procedural persona approach also falls short. For example, the game designer may want to see how different players complete a game with multiple endings. To model these players, she trains an agent that mimics the Exit persona, and she analyzes the trajectory from this agent's execution. Nevertheless, the resultant trajectory of this persona will be the path to the closest ending. The other endings in the game will be neglected, and the game designer will only have playtest data that corresponds to one possible end of the game. A preliminary solution to this problem is masking the feedback from some of the endings. Thus, the agent will generate a playtest towards a particular ending. However, this solution requires additional tinkering, and there might be additional playtests towards the same ending. Another subpar solution is that the game designer would apply randomness to the agent's actions or add random noise to the input to diversify the trajectories. However, randomness does not guarantee that the agent will generate different playtests. Therefore, this solution also does not give complete control to the game designer.

On the other hand, with human playtesters, the game designer could have asked a playtester to play differently. The playtester already knows which paths or particular states she has visited before, so she uses this past knowledge to play the game differently. Therefore, the source of this problem is that the current agent does not know what the previous agents did in the prior runs. Every playtester which an RL agent represents generates a playtest anew. In order to solve this problem, we propose Alternative Path Finder.

\subsubsection{Measuring Similarity}\label{sec:m:ss}

A game can be formulated using a Markov Decision Process (MDP). MDP formulates the interaction between an actor and the environment \cite{Sutton:2018}.

Suppose a human player or an agent played a game, and we obtain the trajectory $\tau{=}$ $\{s_0, a_0, s_1, a_1, ..., s_n\}$ where $s$ corresponds to a state, $a$ corresponds to an action, and the subscripts denote the state or action at time $t$. We want to train an agent that knows $\tau$, and we want this agent to generate a trajectory different than $\tau$. Therefore, we need to calculate a measure to represent the similarity of these two trajectories. We propose two different methods to calculate the similarity. First method is to calculate the recoding probability of a state $s$, $p(s|\tau)$. If $s \in \tau$, then the probability should be high, and if $s \notin \tau$, then the probability should be low. Second method is calculating the prediction error of a dynamics model $q((s_{t}, a_{t}, s_{t+1})|\tau)$. If the transition $s_{t}, a_{t}, s_{t+1}$ exists in $\tau$, then the prediction error should be low, and if this transition does not exist in $\tau$, then the error should be high.

In the rest of this paper, we swap the state $s$ with observation $o$, which the RL agent sees. In most of the frameworks such as GVG-AI~\cite{GVGAI:2019} and VizDoom~\cite{VizDoom:2016}, the observation $o$ seen by the RL agent corresponds to a frame $f$.

\subsubsection{From Recoding Probability to Intrinsic Feedback} \label{sec:m:vpif}

\begin{figure}[t!]
    \centering
    \subcaptionbox{L-shaped Filter\label{fig:m:ss:cts_l}}
        {\includegraphics[width=0.48\columnwidth]{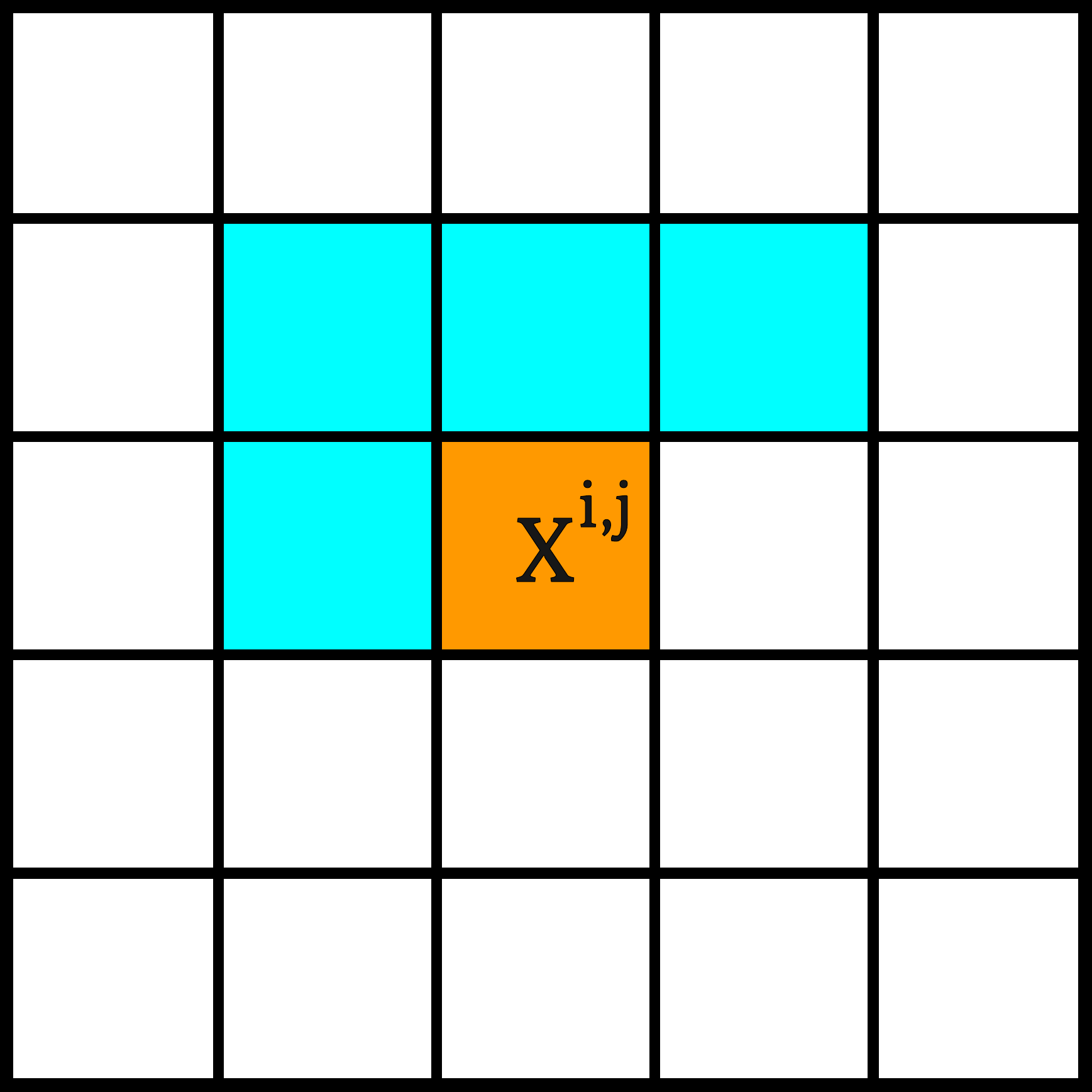}} \hfill
    \subcaptionbox{+-shaped Filter\label{fig:m:ss:cts_p}}
        {\includegraphics[width=0.48\columnwidth]{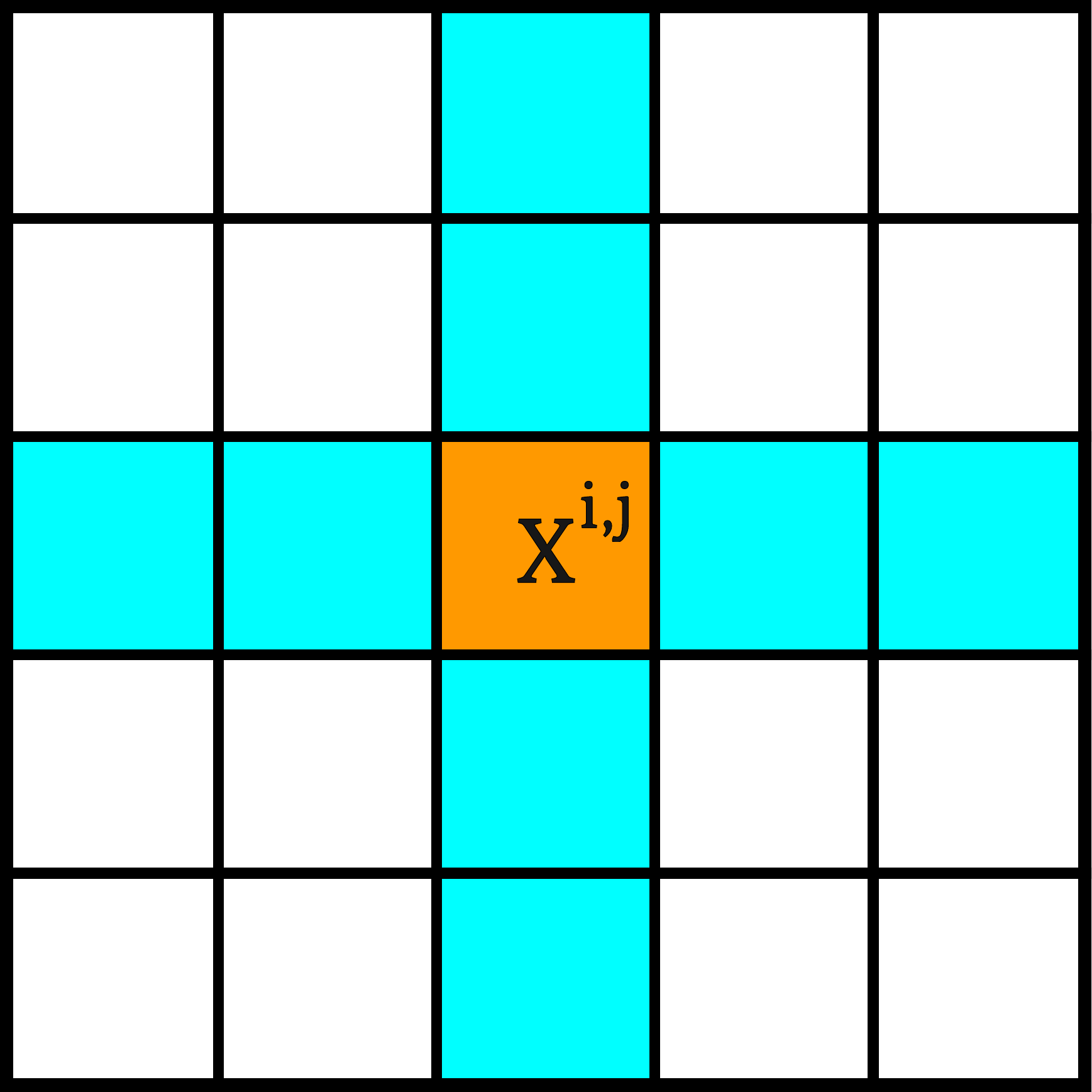}} \hfill
        \caption{Filters mask the pixels around the orange pixel, the data from white pixels are blocked, and the data from the cyan pixels are supplied. Finally, CTS uses the information gathered from cyan pixels to predict the recoding probability of the orange pixel.}
    \label{fig:cts}
\end{figure}

Bellemare et al.~\cite{Bellemare:2016} used Context Tree Switching (CTS)~\cite{Bellemare:2014} to intrinsically motivate an RL agent for exploration. CTS uses a filter to evaluate the recoding probability of a pixel. The filter used by the authors and in our experiments is shown in \figurename{ \ref{fig:m:ss:cts_l}} and \figurename{ \ref{fig:m:ss:cts_p}}, respectively. The filter gathers information around a pixel and CTS uses this information to predict this pixel. When this operation is done for every pixel of an image, the recoding probability of an image is calculated.

In order to use the recoding probability to differentiate between the novel frames from similar frames, we need a boundary probability value. We refer to this probability as $p_{min}$ (see Eq. \ref{eq:cts:p_min}). First, we train a CTS model using all of the frames in trajectories. Then, we calculate the recoding probability of every frame in this trajectory. Next, we set the $p_{min}$ equal to the minimum of all these recoding probabilities. As CTS is a learning-positive model, every frame from these trained trajectories will have a higher recoding probability than $p_{min}$.

\begin{equation}
    \label{eq:cts:p_min}
    \begin{aligned}
        p_{min} = \min(p(f_0|\text{CTS}), p(f_1|\text{CTS}), ..., p(f_n|\text{CTS})) \\
        s.t.~f_{0..n} \in \tau_0, ..., \tau_n
    \end{aligned}
\end{equation}

When an agent or a human player plays the game, the actor will receive a new frame $f_{new}$. First, we calculate its recoding probability $p_{new}{=}p(f_{new}|\text{CTS})$. If $p_{new}$ is smaller than $p_{min}$, this indicates that this frame provides new information and if $p_{new}$ is greater than $p_{min}$, this indicates that this frame does not provide new information. Next, the magnitude of the information depends on how close $p_{new}$ is to $p_{min}$. We use this difference to calculate the amount of reward or penalty.

\begin{equation}
    \label{eq:cts:reward_penalty}
    \begin{aligned}
        p_{new} > p_{min}: feedback = \frac{\beta}{1 + \log\frac{p_{new}}{p_{min}}} - \beta \\
        p_{new} \leq p_{min}: feedback = \beta - \frac{\beta}{1 + \log\frac{p_{min}}{p_{new}}}
    \end{aligned}
\end{equation}
    
We use Eq. \ref{eq:cts:reward_penalty} to calculate the additional reward signal. This formula yields maximum $\beta$ reward when $p_{new} \to 0$ and minimum $-\beta$ when $p_{new} \to 1$. This additional reward signal provides a negative feedback for visiting similar states and positive feedback for visiting novel states. We refer to the APF method that uses CTS internally as APFCTS.

\subsubsection{From Predicting Dynamics to Intrinsic Feedback} \label{sec:m:pdif}

Pathak et al.~\cite{Pathak:2017} used the Intrinsic Curiosity Module (ICM) to intrinsically motivate an RL agent for exploration. ICM is a Neural Network (NN) architecture that learns to predict the environment dynamics and uses the prediction error as the intrinsic motivation. ICM has two NNs called as forward model and inverse model. The forward model predicts the next state features $\phi(s_{t + 1})$ using the current state features $\phi(s_{t})$ and current action $a_t$. The inverse model predicts the current action $a_t$ using the current state features $\phi(s_{t})$ and the next state features $\hat{\phi}(s_{t + 1})$. ICM uses Convolutional Neural Network (CNN) to encode the states into state features, $\phi(s_{t}) = \text{CNN}(s_{t + 1})$. The prediction error is the difference between the predicted next state features $\hat{\phi}(s_{t + 1})$ and extracted next state features $\phi(s_{t + 1})$. Therefore, if the agent has seen the transition $\phi(s_{t}), a_{t}, \phi(s_{t + 1})$, the prediciton error will be low, and if not, the prediction error will be high.

In order to use the prediction error to differentiate between the novel frames from similar frames, we need a boundary value. We refer to this value as $q_{mean}$ (see Eq. \ref{eq:icm:q_mean}). First, we initialize an empty ICM architecture. Next, we use transfer learning to set the weights of CNN encoders, and then we freeze the weights of CNN. The source can be the CNN layers of the RL agent, or if the agent also used ICM, we can use ICM's CNN layers. Afterward, we use the previous trajectories to train the forward and inverse models of ICM. At the end of the training, we have an ICM model that has a better prediction towards the transitions that exist in the given trajectories and a worse prediction towards the transitions that do not exist. Lastly, we replay the previous trajectories, gather all of the prediction errors, and calculate the \textit{mean} of all the prediction errors. We do not calculate the \textit{max} of all the prediction errors as the ICM may not improve the predictions for every transition or make prediction errors. Therefore, \textit{max} would be a poor choice for a boundary value.

\begin{equation}
    \label{eq:icm:q_mean}
    \begin{aligned}
        q_{mean} = \mean(\text{ICM}(f_0, a_0, f_1), ..., \text{ICM}(f_{n-1}, a_{n-1}, f_n)) \\
        s.t.~f_{0..n} \in \tau_0, ..., \tau_n \\
        s.t.~a_{0..n-1} \in \tau_0, ..., \tau_n
    \end{aligned}
\end{equation}

When an agent or a human player plays the game, the actor executes action $a$ on frame $f$. As a result, the actor sees a new frame $f_{new}$. First, we calculate the prediction error of this transition, $q_{new}{=}\text{ICM}(f, a, f_{new})$. If $q_{new}$ is greater than $q_{mean}$, this indicates that this transition is less likely to exist in the previous trajectories. If $q_{new}$ is less than $q_{mean}$, this indicates that this transition is likely to exist in the previous trajectories.

\begin{equation}
    \label{eq:icm:reward_penalty}
    \begin{aligned}
        q_{new} > q_{mean}: feedback = \beta - \frac{\beta}{1 + \log\frac{q_{new}}{q_{mean}}} \\
        q_{new} \leq p_{min}: feedback = \frac{\beta}{1 + \log\frac{q_{mean}}{q_{new}}} - \beta
    \end{aligned}
\end{equation}
    
We use Eq. \ref{eq:icm:reward_penalty} to calculate the additional reward signal. This formula yields maximum $\beta$ reward when $q_{new} \to 0$ and minimum $-\beta$ when $q_{new} \to \infty$. We use this additional feedback signal to reward the novel transitions and to penalize similar transitions. We refer to the APF method that uses ICM internally as APFICM.

\subsubsection{APF Architecture}

\begin{figure}[h!]
    \centering
    \includegraphics[width=0.8\columnwidth]{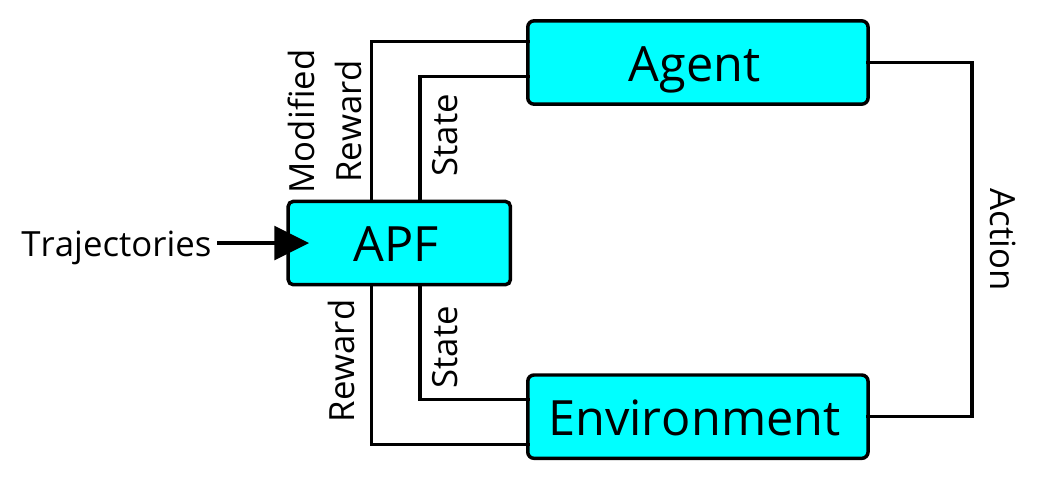}
    \caption{Alternative Path Finding Architecture.}    
    \label{fig:apf_arch}
\end{figure}

We augment the traditional Agent and Environment interaction by adding a new box. This augmented architecture is shown in \figurename{ \ref{fig:apf_arch}}. The APF corresponds to either APFCTS or APFICM. Before we start training an agent, we first train the APF with the previous trajectories as described in Section \ref{sec:m:vpif} or Section \ref{sec:m:pdif}. At this point, we have an APF module that discerns the states or transitions. Afterward, when a new state and a new reward are observed from the environment during the training, these observations first enter the APF. APF modulates the reward signal by adding a penalty or reward by using the Eq. \ref{eq:cts:reward_penalty} or Eq. \ref{eq:icm:reward_penalty}.

The one drawback of this approach is that the feedback is unbounded. Since the feedback is infinite, the agent may loop over novel states or get stuck in a novel state~\cite{Burda:2018}. The agent may visit a novel state repeatedly to get a positive reward and forget the actual task in the environment. The second drawback is that some portion of the game may be strict, offering no alternative paths such as Super Mario Bros.~\cite{Pathak:2017}. Consequently, APF will penalize this portion of the game, naively thinking there may be alternative paths.

We propose a solution for each of these drawbacks. For the first drawback, we propose to put a cap on the total reward and penalty that APF provides. This solution limits the infinite feedback, and this process operates as follows: if a state is distinct, APF clamps the reward by the positive cap $pos_{cap}$. Then, APF yields this clamped reward and updates the positive cap by subtracting the clamped reward. Once the positive cap is exhausted, the additional reward that APF provides becomes zero. We also apply the same principles for the penalty by providing a negative cap, $neg_{cap}$. This solution limits the agent looping over distinct states or getting stuck in a state like the noisy TV problem~\cite{Burda:2018}. Furthermore, as the total reward and penalty are known beforehand, this solution also simplifies the design of the utility function for personas. For the second drawback, we propose to cut these portions from the collected trajectories. Consequently, APF will not penalize the agent, as APF will be blind for this portion of the path.

We introduced two different APF approaches as each has its advantages and disadvantages. The advantage of APFCTS is that the CTS model can be trained from a trajectory that consists of a few frames. However, APFICM is more data-intensive compared to APFCTS. Furthermore, APFICM requires a previously trained agent for transfer learning, which is not required for APFCTS. Nevertheless, as APFCTS operates directly on pixels, a slight noise in a frame would decrease the recoding probability.

Last but not least, though we presented the APF on top of exploration methods CTS and ICM, APF may also be formulated on other exploration methods such as exemplar models~\cite{Fu:2017}. As APF depends on methods used in exploration, we need to draw a line between exploration and APF. The goal of exploration methods is to increase the agent's knowledge about its environment during training. So that when we evaluate, this agent delivers top performance in this environment. The goal of APF is to help the agent to discover the different performances without changing the agent's goal. Therefore during training, APF modulates the reward structure so that the old performances are penalized, and different performances are rewarded.

\section{Experiments}\label{sec:experiments}

In this paper, we used two different environments to test our proposals, GVG-AI~\cite{GVGAI:2019} and VizDoom~\cite{VizDoom:2016}. We describe the environments and the experimental setup in this section.

\begin{figure}[h!]
    \centering
    \includegraphics[width=0.8\columnwidth]{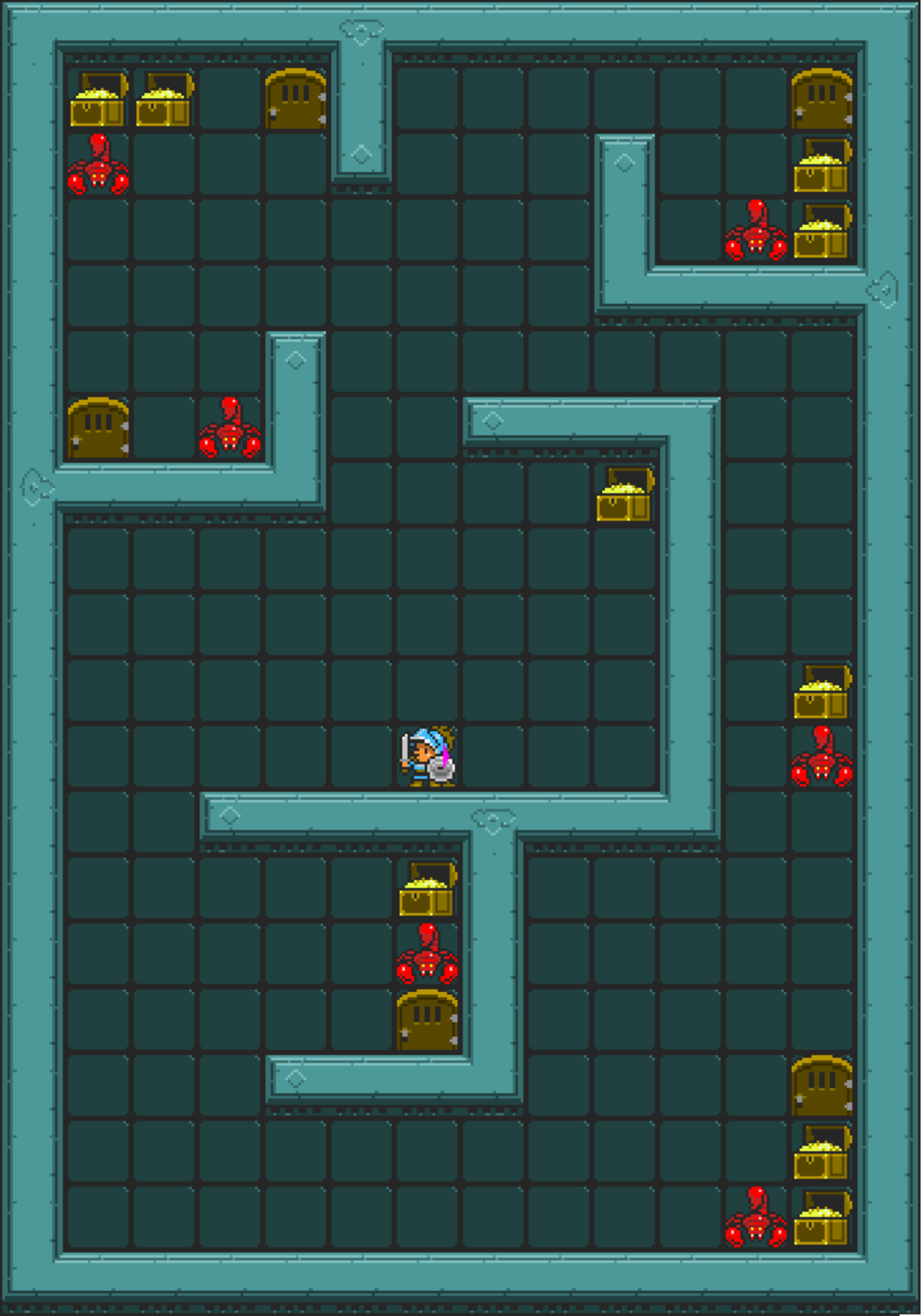}
    \caption{Map of the first testbed game.}
    \label{fig:e:gvgai_game}
\end{figure}

The first testbed game is created using the GVG-AI framework, shown in \figurename{ \ref{fig:e:gvgai_game}}. The game has a $14\times20$ grid-size, and consists of an \textit{Avatar}, \textit{Exits}, static \textit{Monsters}, \textit{Treasures}, and \textit{Walls}. The human player or an agent controls the \textit{Avatar}. The game lasts until the \textit{Avatar} goes to one of the \textit{Exits}, or gets killed by a \textit{Monster}, or until 200 timesteps. The action space consists of six actions \textit{No-Op}, \textit{Attack}, \textit{Left}, \textit{Right}, \textit{Up}, and \textit{Down}. GVG-AI framework is extended to run a game with more than one \textit{Door}. The actor receives distinct feedback for the following interactions killing a \textit{Monster}, getting killed by a \textit{Monster}, collecting a \textit{Treasure}, and colliding with a \textit{Door}.

\begin{figure}[h!]
    \centering
    \includegraphics[width=0.9\columnwidth]{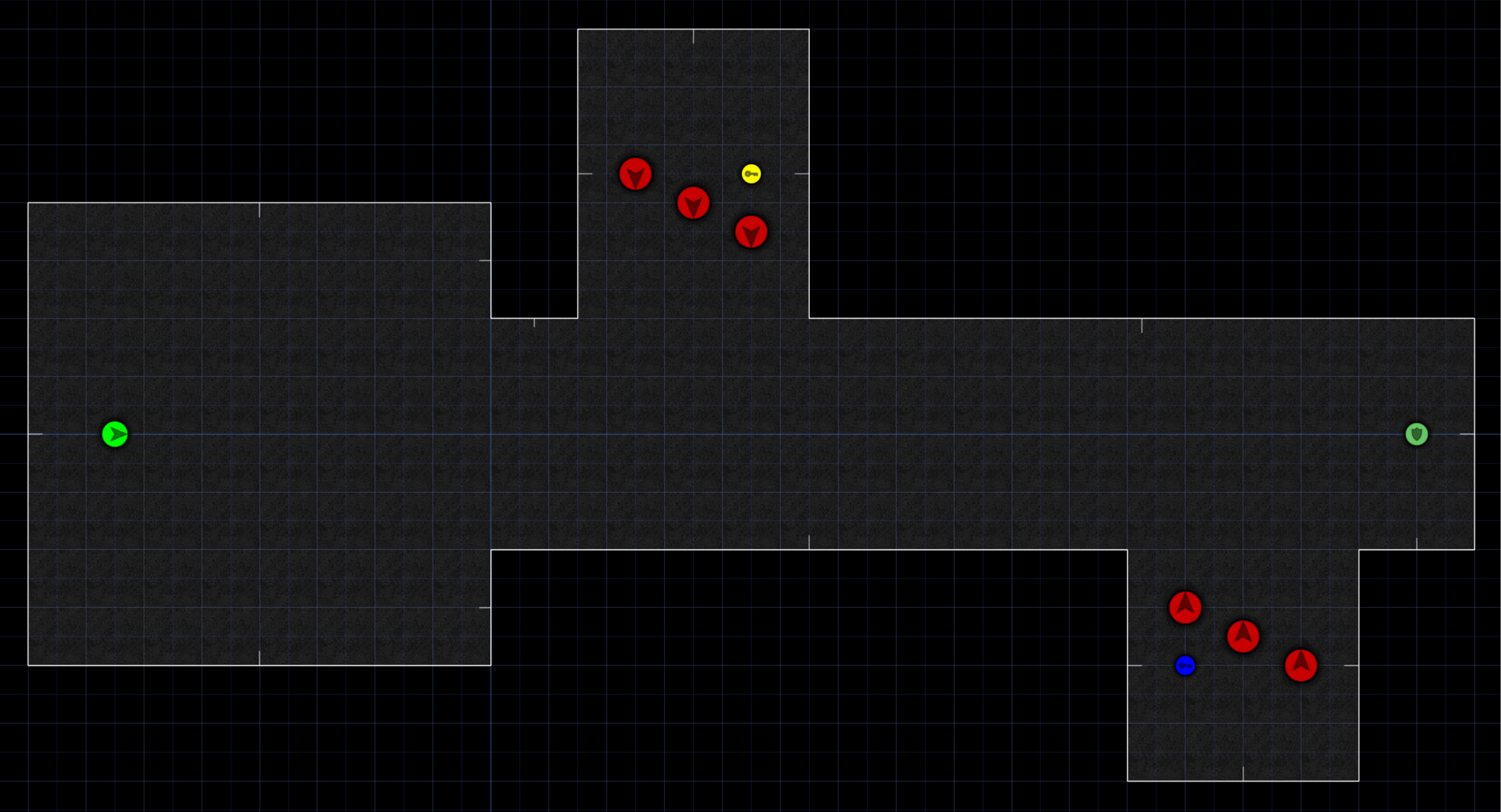}
    \caption{Map of second testbed game.}
    \label{fig:e:first_game}
\end{figure}

The second testbed game is a Doom level, shown in \figurename{ \ref{fig:e:first_game}}. The game has a $1600\times832$ grid size, and consists of an \textit{Avatar}, \textit{Exit}, \textit{Monsters}, \textit{Treasures}, and \textit{Walls}. The human player or an agent controls the \textit{Avatar}. The game lasts until the \textit{Avatar} goes to the \textit{Door}, or gets killed by a \textit{Monster}, or until 2000 timesteps. The action space consists of seven actions \textit{Attack}, \textit{Move Left}, \textit{Move Right}, \textit{Move Up}, \textit{Move Down}, \textit{Turn Left}, and \textit{Turn Right}. The actor receives distinct feedback for the following interactions killing a \textit{Monster}, getting killed by a \textit{Monster}, collecting a \textit{Treasure}, and colliding with the \textit{Door}. Additionally, the actor receives constant negative feedback of $0.001$ for every step taken.

\begin{figure}[h]
    \centering
    \includegraphics[width=0.8\columnwidth]{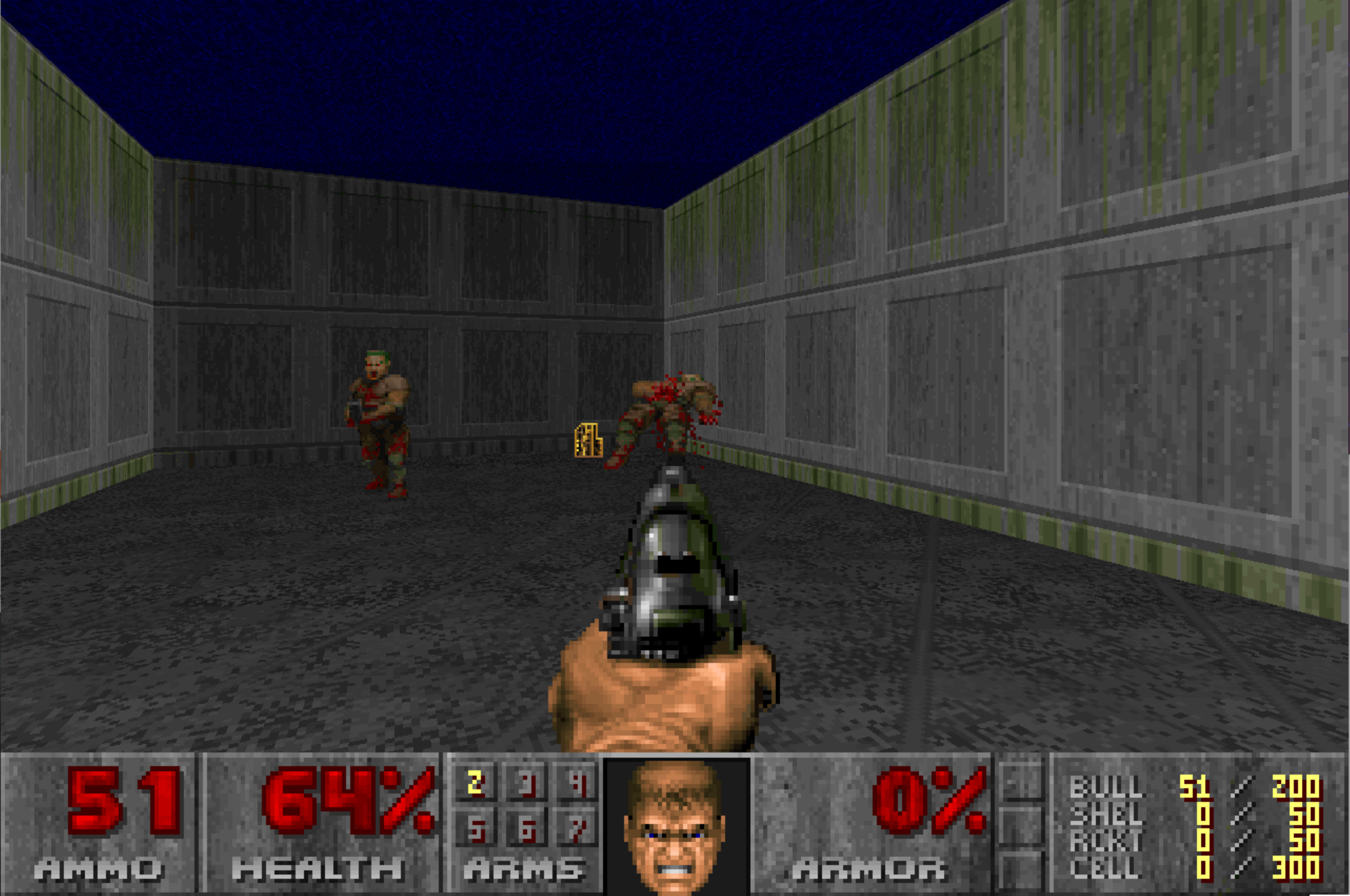}
    \caption{Doom in-game snapshot.}
    \label{fig:e:doom_in_game}
\end{figure}

\begin{figure}[h!]
    \centering
    \includegraphics[width=0.9\columnwidth]{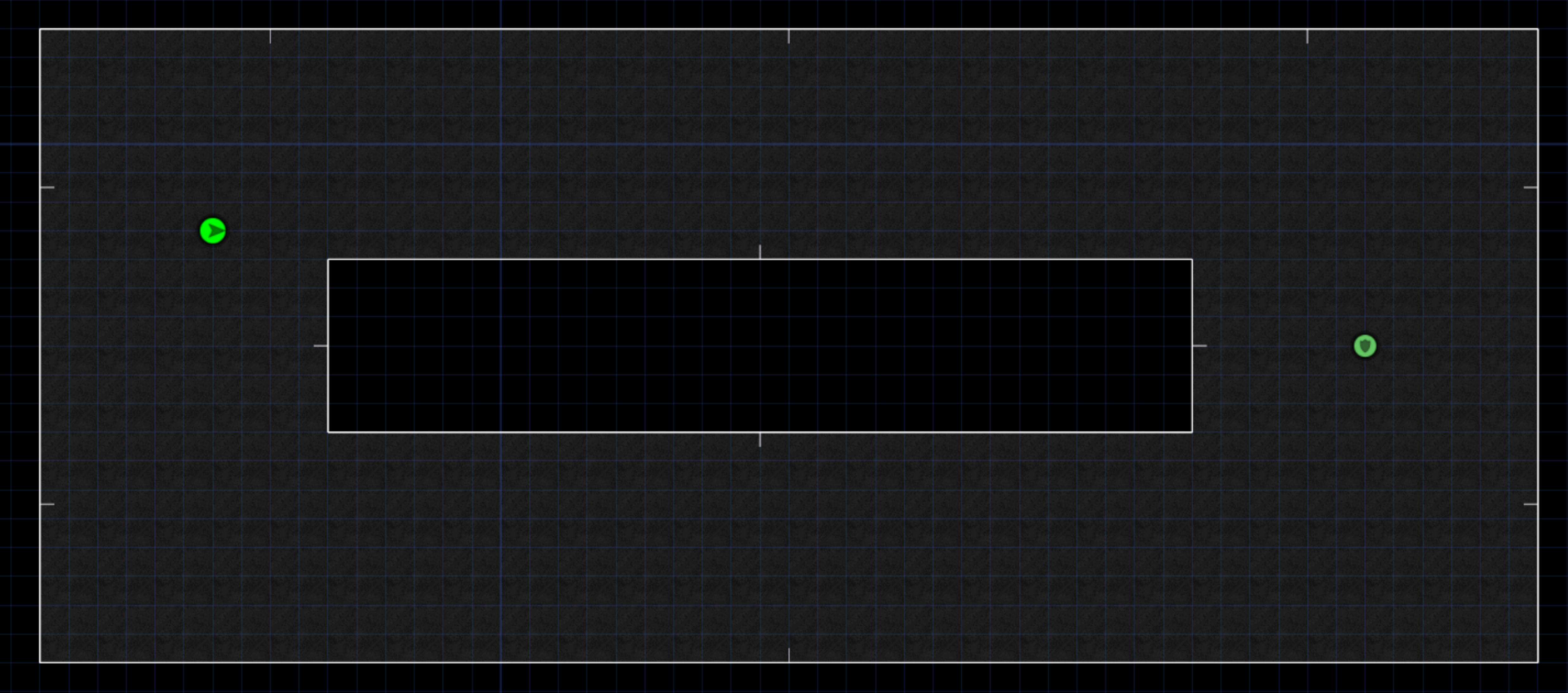}
    \caption{Map of third testbed game.}
    \label{fig:e:second_game}
\end{figure}

The third testbed game is another Doom level, shown in \figurename{ \ref{fig:e:second_game}}. The game has a $1664\times704$ grid size, and consists of an \textit{Avatar}, an \textit{Exit}, and \textit{Walls}. The human player or an agent controls the \textit{Avatar}. The game lasts until the \textit{Avatar} goes to the \textit{Door}, or until 2000 timesteps. The action space consists of seven actions \textit{Attack}, \textit{Move Left}, \textit{Move Right}, \textit{Move Up}, \textit{Move Down}, \textit{Turn Left}, and \textit{Turn Right}. The actor receives feedback if the actor interacts with the \textit{Door}. Additionally, the actor receives constant negative feedback of $0.001$ for every step taken.

We experiment with the procedural and goal-based personas in the first and second testbed games. We test the APF in the first and third testbed games. We used the same random seed during the APF experiment to properly test the APF method. We use PPO~\cite{Schulman:2017} agent in all of the experiments. For the PPO+CTS, PPO+ICM, PPO+APFCTS, and PPO+ICM+APFICM, we change the base PPO implementation slightly. The base PPO implementation is from the Stable-Baselines project~\cite{stable-baselines}. We also tested the proposed persona with other RL agents during the initial experiments, and we found that PPO requires less hyperparameter tuning, so we used PPO in all of our experiments. The hyperparameters of PPO agents are presented in Table \ref{table:e:agents}, and the hyperparameters of APF techniques are shown in Table \ref{table:e:apf}. Lastly, as the first game is deterministic, we evaluated the trained agent once. On the other hand, as the second and the third games are stochastic, we evaluated the trained agent 1000 times. Furthermore, we noticed that our training was more consistent whenever we used an exploration algorithm such as CTS or ICM. Consequently, we had to restart the training in the first game.

GVG-AI environment sends an observation with shape $160 \times 112 \times 4$, we downscale this observation to $80 \times 56$ and then convert the observation into grayscale. Afterward, we stack the most recent four observations, and lastly feed the stacked observations to the agent. For CTS used in PPO+CTS and APFCTS, we process the observation into we $42 \times 42$, 3-bit grayscale image, and calculate the recoding probability of this observation. Doom environment sends the observation with shape $160 \times 120 \times 1$, we resize this observation to $84 \times 84 \times 1$, and we feed the agent and the APFICM this resized observation.

We created four different procedural personas and five different developing personas. The four procedural personas are Exit, Monster Killer, Treasure Collector, and Completionist. The utility weights of these procedural personas is given in Table~\ref{table:e:procedural_personas}. We chose these procedural personas from \cite{Holmgard:2018}, and we drew inspiration from these personas to make their developing persona counterparts. The five developing personas are Developing Monster Killer, Developing Treasure Collector, Developing Raider, Developing Completionist, and Developing Casual Completionist. The development sequences of these personas are presented in Table~\ref{table:e:developing_persona}, the utility function of the goals are given in Table~\ref{table:e:utility_goals}, and the criteria of these goals are shown in Table~\ref{table:e:goal_criteria}.

\renewcommand{\arraystretch}{1.2}
\begin{table}[h!]
\centering
\captionsetup{justification=centering}
\caption{Utility weights for procedural personas. Exit (E), Monster Killer (MK), Treasure Collecter (TC), and Completionist (C).}
\begin{tabular}{ccccc}
&  \multicolumn{4}{c}{ \textbf{Personas}}\\ 
\cline{2-5} \\
\textbf{Game Event}    & (E)    & (MK)    &  (TC)   & (C)  \\
\hline
Reaching an Exit       &  ~1    &   ~0.5  &  ~0.5   &      \\
Killing a Monster      &        &   ~1    &         & ~1   \\
Collecting a Treasure  &        &         &  ~1     & ~1   \\
Dying                  &  -1    &   -1    &  -1     & -1   \\
\end{tabular}
\label{table:e:procedural_personas}
\end{table}

\renewcommand{\arraystretch}{1.2}
\begin{table}[h!]
\centering
\captionsetup{justification=centering}
\caption{Utility weights for the goals. Killer (K), Collecter (Col), Exit (E), and Completionist (Com).}
\begin{tabular}{ccccc}
&  \multicolumn{4}{c}{ \textbf{Goal Names}}\\ 
\cline{2-5} \\
\textbf{Game Event}    & (K)    & (Col)     & (E)   & (Com) \\
\hline
Death                  &  -1    &  -1       &  -1   &     -1        \\
Exit Door              &        &           &  ~1   &               \\
Monster Killed         &  ~1    &           &       &     ~1        \\
Treasure Collected     &        &  ~1       &       &     ~1        \\
\end{tabular}
\label{table:e:utility_goals}
\end{table}

\renewcommand{\arraystretch}{1.2}
\begin{table}[h!]
\centering
\captionsetup{justification=centering}
\caption{Sequences for the developing personas.}
\begin{tabular}{cl}
\textbf{Hyperparameters}    & \multicolumn{1}{c}{\textbf{Development Sequence}} \\
\hline
Dev. Killer                &   Killer -> Exit               \\
Dev. Collector             &   Collector -> Exit            \\
Dev. Raider                &   Killer -> Collector -> Exit  \\
Dev. Completionist         &   Completionist -> Exit        \\
Dev. Casual Completionist  &   Casual Completionist -> Exit \\
\end{tabular}
\label{table:e:developing_persona}
\end{table}

\renewcommand{\arraystretch}{1.2}
\begin{table}[h!]
\centering
\captionsetup{justification=centering}
\caption{Criteria of the goals. Killer (K), Collecter (Col), Completionist (Com), and Casual Completionist (Cas. Com.).}
\begin{tabular}{ccccc}
&  \multicolumn{4}{c}{ \textbf{Goal Names}}\\ 
\cline{2-5} \\
\textbf{Criterion}     & (K)     & (Col)    & (Com)   & (Cas. Com.) \\
\hline
Monsters Killed        &   50\%  &          &   100\%  &             \\
Treasure Collected     &         &  50\%    &   100\%  &             \\
Remaining Health       &         &          &          &   50\%      \\   
\end{tabular}
\label{table:e:goal_criteria}
\end{table}

\vspace{10em}

\section{Results}\label{sec:results}

In this study, we asked the following research questions.
\begin{itemize}
    \item How does a goal-based persona perform compared to a procedural persona?
    \begin{itemize}
        \item Diversity of playtests generated by personas
        \item Agreement between interactions performed and Persona's decision model
    \end{itemize}
    \item Which additional paths can be discovered with APF?
\end{itemize}

\subsection{Experiment I: Procedural vs Goal-based personas:}

Table \ref{table:r:e_1_ppo_results} presents the interactions done by seven different personas. The Exit persona directly goes to the \textit{Door}, which is four spaces below the \textit{Avatar}. The other three procedural personas also go to the same \textit{Door}, but also collecting the \textit{Treasure} and killing the \textit{Monster} on the way. The Developing Killer persona defeats all of the \textit{Monsters} on the upper half of the level. The Developing Collector persona collects four of the \textit{Treasures} on the upper half of the level. The Developing Raider is a combination of Developing Killer and Developing Collector, consequently kills the \textit{Monsters} and then collects the \textit{Treasures} in the upper half of the level. Lastly, the Developing Completionist kills more \textit{Monsters} and collects more \textit{Treasures} than every other persona. However, Developing Completionist misses the \textit{Monster} and the \textit{Treasure} below the starting position. We see all procedural personas interact with a small region of the level, whereas the developing personas interact with a broader region. Therefore, we conducted the same experiment for procedural personas with PPO + CTS RL agent. Table \ref{table:r:e_1_ppo_results} displays the interactions performed by procedural personas when the agent explores the environment. We see that the interactions performed by PPO + CTS RL agent fit better to the persona's decision model.

\renewcommand{\arraystretch}{1.2}
\begin{table}[h!]
\centering
\captionsetup{justification=centering}
\caption{Interactions of Personas performed by the PPO RL agent in Experiment I.}
\begin{tabular}{cccc}
&  \multicolumn{3}{c}{ \textbf{Game Event}}\\ 
\cline{2-4} \\
\textbf{Personas}       & Monsters Killed & Treasures Collected & Door    \\
\hline
Exit                    &       0        &         0          &   1       \\
Monster Killer          &       1        &         1          &   1       \\
Treasure Collector      &       1        &         1          &   1       \\   
Completionist           &       1        &         1          &   1       \\
Dev. Killer             &       3        &         0          &   1       \\
Dev. Collector          &       1        &         4          &   1       \\
Dev. Raider             &       3        &         4          &   1       \\ 
Dev. Completionist      &       5        &         8          &   1       \\   
\end{tabular}
\label{table:r:e_1_ppo_results}
\end{table}

\renewcommand{\arraystretch}{1.2}
\begin{table}[h!]
\centering
\captionsetup{justification=centering}
\caption{Interactions of Personas performed by the PPO + CTS RL agent in Experiment I.}
\begin{tabular}{cccc}
&  \multicolumn{3}{c}{ \textbf{Game Event}}\\ 
\cline{2-4} \\
\textbf{Personas}       & Monsters Killed & Treasures Collected & Door    \\
\hline
Monster Killer          &       2        &         0          &   1       \\
Treasure Collector      &       0        &         3          &   1       \\   
Completionist           &       2        &         3          &   1       \\
\end{tabular}
\label{table:r:e_1_ppocts_results}
\end{table}

\subsection{Experiment II: Alternative paths found in GVG-AI:}

\begin{figure}[h!]
    \centering
    \includegraphics[width=0.8\columnwidth]{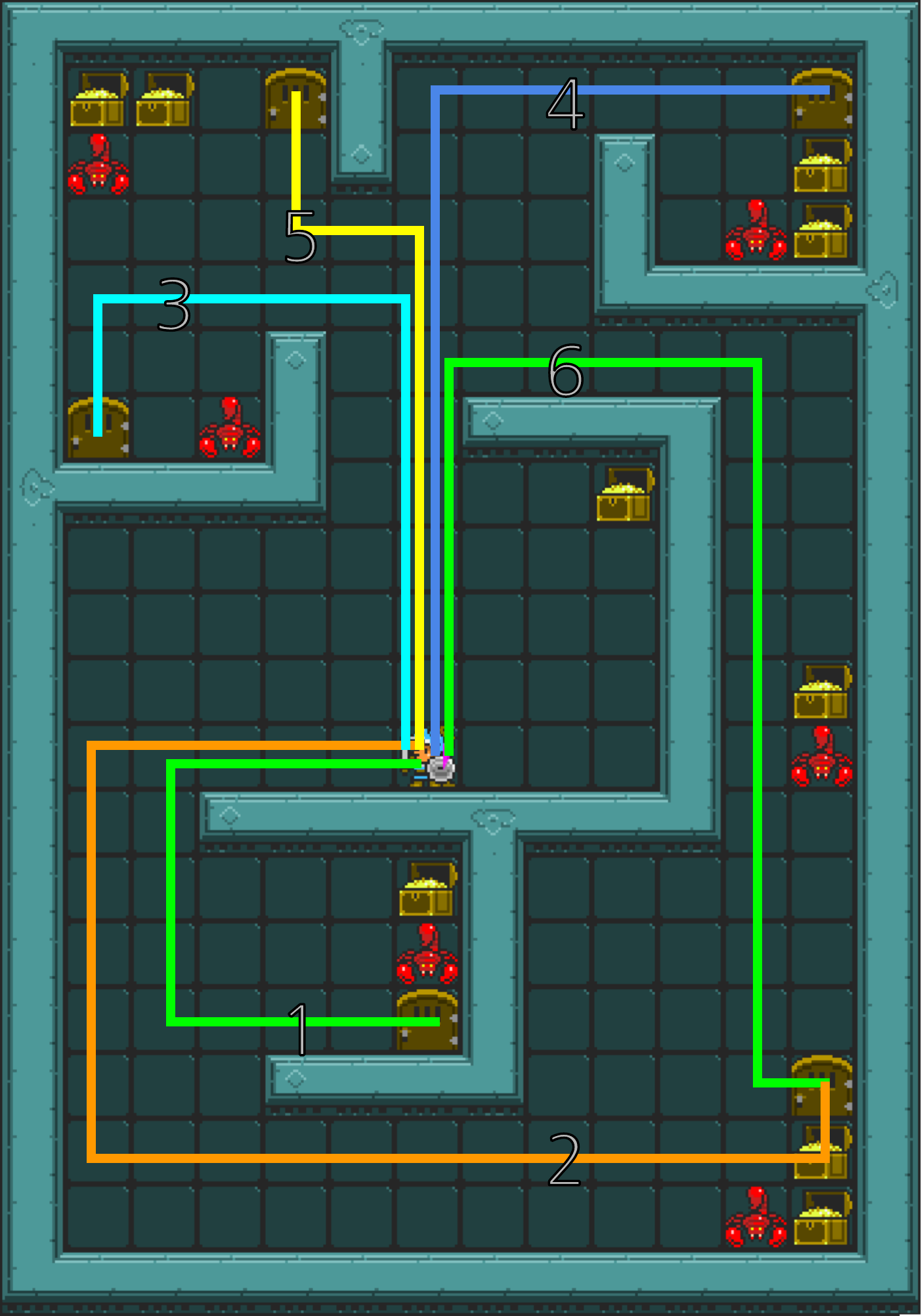}
    \caption{Paths found by Exit persona with PPO and with PPO + CTS + APFCTS.}
    \label{fig:r:gvgai_exit_paths}
\end{figure}

We used the path found by the Exit persona in Experiment I to train APFCTS (see Path 1 in \figurename{ \ref{fig:r:gvgai_exit_paths}}). Then, we trained the PPO + CTS + APFCTS agent in the first testbed game while using the Exit persona's utility weights. We repeated the experiment for each path obtained from the PPO + CTS + APFCTS agent. First, an APFCTS is trained using one of the obtained paths, and then we use this trained APFCTS to train a PPO + CTS + APFCTS agent. The paths identified at the end of the process are shown in \figurename{ \ref{fig:r:gvgai_exit_paths}}. Table \ref{table:r:e_2_results} shows the total discounted rewards---the rewards received from the environment and the APFCTS. The bold values indicate the alternative paths of the trained path. For example, Path 1 has four alternative paths---Paths 2 to 6. Table \ref{table:r:e_2_results} also shows that, when we use APFCTS, we see that the reward of playing the same path decreases by at least $0.1$, and the reward of space-disjoint paths increases by at least $0.1$. This reward difference justifies why APF supports finding alternative paths.

Lastly, from Table \ref{table:r:e_2_results} we notice that APFCTS clusters the paths in Experiment II into two equivalence classes, which are $\{1,2\}$ and $\{3,4,5,6\}$. Therefore, we may interpret that distinct paths refer to \textit{paths that are space-disjoint from the one trained on} for APFCTS.

\renewcommand{\arraystretch}{1.2}
\begin{table}[h!]
\centering
\captionsetup{justification=centering}
\caption{Total Discounted Reward without APFCTS and with APFCTS. The first row shows the total discounted reward without APFCTS. For the rows with a path number, the number indicates which path we used to train the APFCTS. The values under tested paths show the total discounted reward that the agent receives when APFCTS modulates the environment reward. The bold values demonstrate the found paths when we execute the PPO + CTS + APFCTS agent.}
\begin{tabular}{ccccccc}
&  \multicolumn{6}{c}{ \textbf{Tested Paths}}\\ 
\cline{2-7}
\textbf{Trained Path} & 1 & 2 & 3 & 4 & 5 & 6 \\
- &\textbf{0.86}&           0.78  &   0.84       &      0.84         &\textbf{0.86}      &     0.76    \\
1 &  0.76       &           0.77  &\textbf{0.98} & \textbf{0.98}     &\textbf{0.98}      &\textbf{0.98}\\
2 &  0.82       &           0.61  &\textbf{0.98} & \textbf{0.99}     &\textbf{0.98}      &\textbf{0.98}\\
3 &\textbf{0.98}&  \textbf{0.98}  &       0.74   &       0.86        &        0.82       &        0.88 \\
4 &\textbf{0.99}&  \textbf{0.98}  &       0.86   &       0.72        &        0.86       &        0.86 \\
5 &\textbf{0.98}&  \textbf{0.98}  &       0.81   &       0.85        &        0.76       &        0.87 \\
6 &\textbf{0.99}&  \textbf{0.98}  &       0.87   &       0.87        &        0.88       &        0.60 \\
\label{table:r:e_2_results}
\end{tabular}
\end{table}

\subsection{Experiment III: Personas in Doom:}

We experimented with 9 different personas in the second testbed game, a Doom level (see \figurename{ \ref{fig:e:first_game}}). The interaction results are presented in Table \ref{table:r:e_3_results}, and all of the personas behave similarly to their specifications. The Exit persona always finishes the game, and in some of the evaluations, Exit persona kills a \textit{Monster} but never collects a \textit{Treasure}. The Monster Killer persona generally kills all of the \textit{Monsters}, rarely collects a \textit{Treasure}, and habitually finishes the game. Developing Killer is similar to Monster Killer but kills half of the \textit{Monsters} and rarely dies. The Treasure Collector and Developing Collector are alike. They both collect a single \textit{Treasure}, kill the least \textit{Monsters} and die the most. The Completionist, Developing Completionist, and Developing Casual Completionist personas behave similarly, but minor differences exist. The Developing Casual Completionist always finishes the level but usually cannot collect the second \textit{Treasure}. The Completionist and Developing Completionist regularly collect the second \textit{Treasure}, but in doing so, rarely die and cannot finish the level.

\renewcommand{\arraystretch}{1.2}
\begin{table}[h!]
\scriptsize
\centering
\captionsetup{justification=centering}
\caption{Interactions of Personas in Experiment III over 1000 evaluations.}
\begin{tabular}{ccccc}
&  \multicolumn{4}{c}{ \textbf{Game Event}}\\ 
\cline{2-5} \\
\textbf{Personas}       & Monsters         & Treasures       & Door            & Death \\
\hline
Exit                    &  0.27 $\pm$ 0.48 & 0.00 $\pm$ 0.00 & 1.00 $\pm$ 0.00 & 0.00 $\pm$ 0.00\\
MK                      &  5.79 $\pm$ 0.91 & 0.01 $\pm$ 0.07 & 0.98 $\pm$ 0.15 & 0.00 $\pm$ 0.00\\
Dev. Killer             &  3.54 $\pm$ 0.98 & 0.01 $\pm$ 0.08 & 0.96 $\pm$ 0.19 & 0.01 $\pm$ 0.08\\
TC                      &  1.94 $\pm$ 0.70 & 0.94 $\pm$ 0.24 & 0.80 $\pm$ 0.40 & 0.19 $\pm$ 0.39\\
Dev. Collector          &  2.00 $\pm$ 0.65 & 0.95 $\pm$ 0.22 & 0.87 $\pm$ 0.34 & 0.13 $\pm$ 0.34\\
Dev. Raider             &  3.52 $\pm$ 0.73 & 0.98 $\pm$ 0.15 & 0.97 $\pm$ 0.17 & 0.01 $\pm$ 0.08\\
Comp.                   &  5.76 $\pm$ 1.06 & 1.91 $\pm$ 0.38 & 0.95 $\pm$ 0.22 & 0.01 $\pm$ 0.11\\
Dev. Comp.              &  5.81 $\pm$ 0.92 & 1.91 $\pm$ 0.36 & 0.96 $\pm$ 0.19 & 0.01 $\pm$ 0.09\\
Dev. Cas. Comp.         &  5.83 $\pm$ 0.53 & 0.98 $\pm$ 0.13 & 0.98 $\pm$ 0.14 & 0.00 $\pm$ 0.00\\
\end{tabular}
\label{table:r:e_3_results}
\end{table}

\subsection{Experiment IV: Alternative paths found in Doom:}

\begin{figure}[h!]
    \centering
    \includegraphics[width=0.9\columnwidth]{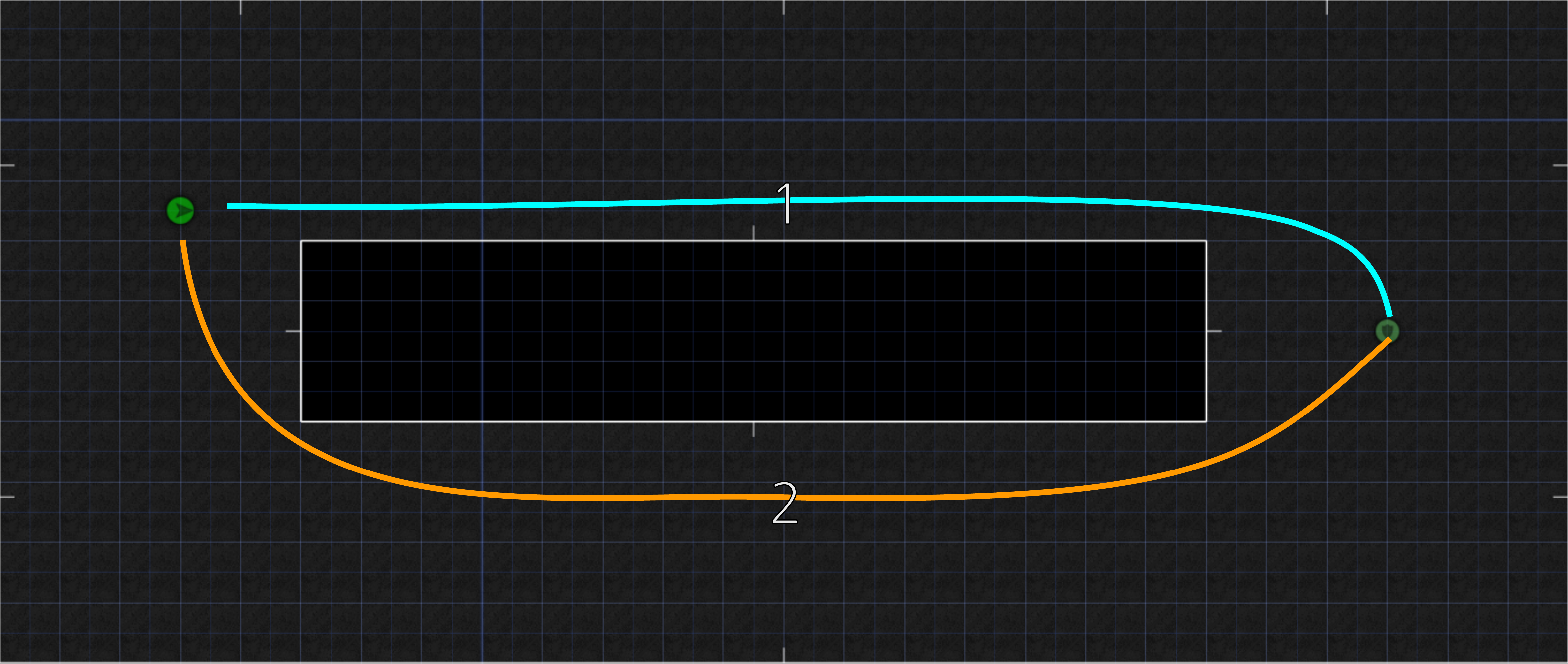}
    \caption{Paths found by Exit persona with PPO and with PPO + ICM + APFICM.}
    \label{fig:r:e_4_results}
\end{figure}

We trained an Exit persona in the third testbed game using PPO + ICM agent. The first path shown in \figurename{ \ref{fig:r:e_4_results}} is the trajectory taken by the Exit persona. We trained an APFICM using this first path, and then we trained a new Exit persona using PPO + ICM + APFICM agent. The new Exit persona played the second path. The total discounted reward obtained by these two Exit personas is shown in Table \ref{table:r:e_4_results}. As the first path consists of 52 steps, whereas the second path consists of 77 steps, the total reward of the first path is higher than the second. However, applying APFICM, we increase the total reward obtained from the second path and decrease the total reward obtained from the first path.

\renewcommand{\arraystretch}{1.2}
\begin{table}[h!]
\centering
\captionsetup{justification=centering}
\caption{Total Discounted Reward without APFICM and with APFICM over 1000 evaluations. First row shows the total discounted reward without APFICM. For the rows with a number, we train the APFICM and calculate the discounted reward by using ICM. The bold values demonstrate the found paths when we execute the PPO + ICM + APFICM agent by training APFICM with the trained path.}
\begin{tabular}{ccc}
&  \multicolumn{2}{c}{ \textbf{Tested Paths}}\\ 
\cline{2-3}
\textbf{Trained Path} & 1                        & 2                         \\
-                     &\textbf{0.80 $\pm$ 0.02} & 0.68 $\pm$ 0.01 \\
1                     & 0.51 $\pm$ 0.02         & \textbf{0.78 $\pm$ 0.01} \\
\label{table:r:e_4_results}
\end{tabular}
\end{table}

\section{Discussion}\label{sec:discussion}

In this paper, we presented an advancement for procedural persona, goal-based persona and introduced a method to let RL agents discover different paths, APF. We experimented with these methods in GVG-AI and Doom environments.

Procedural personas and developing personas are two methods used by game designers to automate the playtesting process. One drawback of the procedural personas originates from the utility function. A utility function realizes the decision model of a persona. For example, a Treasure Collector receives positive feedback from finishing the level and collecting a \textit{Treasure}. However, if the starting position of the agent is close to the \textit{Door}, the agent may neglect the \textit{Treasures}. Conversely, if the \textit{Door} is positioned after the \textit{Treasures}, the agent is likely to interact with most of the \textit{Treasures}. We saw this dilemma in Experiment I. Without any exploration technique, the procedural personas Monster Killer, Treasure Collector, and Completionist executed the same set of actions. When we integrated exploration into the agents that realize these personas, the set of actions executed by these personas became different. Furthermore, these new sets of actions were more fitting to their decision model. This problem is also seen in the MCTS agent playtesting the MiniDungeons 2 game \cite{Holmgard:2018}. 

The problem with the utility function is that the utility function is an amalgamation of multiple goals. Hence, depending on the level composition and RL agent's hyperparameters, the procedural persona represents one of those playstyles. In Experiment I, we believe that the Developing Completionist fits better with the idea of a ``Completionist'' persona than the procedural Completionist. Developing persona addresses this problem by introducing a sequence of goals. Consequently, a game designer may use the developing persona to choose which playstyle she wants to playtest carefully.

Another advantage of developing personas over procedural personas is that developing personas support playstyles that involve alteration. For example, in Experiment I, the Developing Raider killed the \textit{Monsters} and then collected the \textit{Treasures}. The Developing Raider starts the game as a Monster Killer and becomes a different persona ---a Treasure Collector--- after fulfilling a criterion. These development sequences were mentioned by Bartle~\cite{Bartle:1996}, but development sequences were impractical while using a single utility function. Consequently, this behavior performed by Developing Raider was missing in procedural personas. On the other hand, another important aspect of playtesting is the ability to generate playtraces as if a human would. In this paper, we used handcrafted utility functions, however, these utility functions could have been extracted from human playtest data by Inverse Reinforcement Learning~\cite{Sutton:2018}. This alteration might help the RL agent to generate a playtest that is more human-like~\cite{Ariyurek:2019}\cite{Tastan:2011}.


In addition to the GVG-AI environment, we conducted experiments on the Doom environment. To the best of our knowledge, this paper is the first study to playtest personas in a 3D environment. In 2D environments, the researchers~\cite{Holmgard:2018}\cite{Mugrai:2019} employed MCTS RL agent to realize personas. Nevertheless, MCTS would be an ineffective choice for 3D environments, and MCTS would underrepresent the persona. Consequently, we used the PPO agent in Experiments III and IV, as PPO is a competent agent used by OpenAI~\cite{Baker:2020}. In Experiment III, we see that the PPO agent realized the decision models of personas properly. From the results in Table \ref{table:r:e_3_results}, we interpret that a player has to kill a \textit{Monster} to finish this level. The level is hardest for Treasure Collector and Developing Collector as they have to kill a \textit{Monster} to collect the \textit{Treasures}. We see an interesting fact about the game when we compare the Developing Casual Completionist and the Developing Completionist personas. The former never dies but collects only a single \textit{Treasure}, whereas the latter seldom dies but collects both of the \textit{Treasures}. From this data, we understand that collecting the second \textit{Treasure} causes the death of the player. As the Developing Casual Completionist fears losing her health more, this persona finds collecting the second \textit{Treasure} risky. Furthermore, comparing the Killer and the Developing Killer personas shows that the latter die more than the former. This comparison unravels another fact about this level. If a player engages in combat to kill \textit{Monsters}, then this player should kill as much as possible. Otherwise, this player is likely to die, such as the Developing Killer. On the other hand, the Developing Casual Completionist also kills as much \textit{Monsters} as a Monster Killer. This indifference indicates that the game may not be challenging enough for a hardcore player.

In Experiment II, we prepared a game that consists of five \textit{Doors}. We found that ---without APF--- the Exit persona would take either the first or the fifth path shown in \figurename{ \ref{fig:r:gvgai_exit_paths}}. The lengths of these paths are the same and shorter than every possible path that ends with a \textit{Door}. Consequently, in the first row of Table \ref{table:r:e_2_results}, we see that the first and the fifth path share the highest score. Furthermore, in Experiment IV, we saw that ---without APF--- the Exit persona would take the first path (see \figurename{ \ref{fig:r:e_4_results}}). Since this path is the closest towards the \textit{Door}, and therefore, playing this path yields a higher score compared to the other path, shown in Table \ref{table:r:e_4_results}.

We proposed APF to let RL agents discover these additional paths shown in \figurename{ \ref{fig:r:gvgai_exit_paths}} and \figurename{ \ref{fig:r:e_4_results}}. A human playtester would have played these paths, but without APF, the Exit persona would overlook them as these paths yield a lower score. Hence, the game designer would not have any playtest data for other endings. Table \ref{table:r:e_2_results} and Table \ref{table:r:e_4_results} show insight on how APF achieves this feat. APF modulates the reward signal of the environment. When the agent tries to learn a similar path, the agent is penalized, and when the agent tries to learn a distinct path, the agent is rewarded. This reward modulation is the reason how APF promotes finding distinct paths. The game designer can exercise the APF to get a distinct path and then study this path to improve her game. Afterward, she can exercise the APF to generate as many paths as she needs. However, the game designer might be interested in examining the play traces that could have come from human playtesters. We could employ an auxiliary NN trained to select the best human-like action given an observation~\cite{Gudmundsson:2018}. Nevertheless, carefully combining this NN with APF is a topic of another study.

On the other hand, an alternative path is a subjective concept. Every human playtester may think of another way to represent the Exit persona. In Table \ref{table:r:e_2_results}, we see that when we train the APFCTS with the second path, the score of the first path decreases, and the score of the sixth path increases. According to APFCTS, the first and second paths are more similar than the second and sixth paths (see \figurename{ \ref{fig:r:gvgai_exit_paths}}). However, one might argue that the first and second paths are distinct as they reach different \textit{Doors}, and the second and sixth paths are similar as they reach the same \textit{Door}. Though APFCTS is objective in finding alternative paths, these alternative paths are ``subjectively'' different for the game designers. The objectivity of APFCTS and APFICM comes from the recoding probability of a frame and the dynamics prediction error, respectively.

Additionally, we found that APFICM is more robust compared to APFCTS. We also experimented with APFCTS in Doom. However, CTS calculated the recoding probability of some frames as $0$. Furthermore, we observed that for our experimentation setup the plus-shaped filter in \figurename{ \ref{fig:m:ss:cts_p}} yielded better results than the original CTS filter in \figurename{ \ref{fig:m:ss:cts_l}}. Lastly, researchers employed curiosity to increase the playtesting coverage of an RL agent~\cite{Gordillo:2021}. Though we promoted APF to find distinct paths, APF may help game tester agents~\cite{Ariyurek:2019}. Coverage is crucial for testing, and APF increases coverage by finding distinct paths.

\textbf{Limitations \& Challenges:} The performance of developing and procedural persona is dependent on the RL algorithms. If the RL algorithm cannot play a game, the game designer could not benefit from these automated playtesters. Furthermore, our APF proposals are based on exploration algorithms. The performance of APF in an environment is linked to how well the exploration algorithm would perform in this environment.

\section{Conclusion}\label{sec:conclusion}

This paper focused on the problem of providing additional tools to game designers for playtesting. In this regard, we proposed developing persona, a direct successor to procedural personas. Furthermore, we presented a novel method to help RL agents to discover alternative trajectories, APF. We introduced two APF approaches, APFCTS and APFICM.

Our results show that developing personas are a successor of procedural personas. A game designer can embody various personalities in developing personas to generate unique playtests. Furthermore, our experiments indicate that developing personas provide information to game designers that procedural personas cannot provide. Furthermore, we show that automated playtesting can be extended to 3D environments using state-of-the-art RL algorithms.

We proposed APF to discover alternative paths in an environment. We based APF on exploration research techniques and proposed two methodologies to implement APF, APFCTS, and APFICM. In our experiments in GVG-AI and Doom environments, we found that APF ensures that the same path is not generated again.

In the future, we would like to experiment with different personas using APF. Next, APFICM can be improved by substituting the linear layer with an LSTM layer. This substitution will provide path information rather than state transition information. Lastly, we would like to experiment with other 3D environments such as Minecraft~\cite{Minecraft:2016}.


\bibliographystyle{IEEEtran}
\bibliography{IEEEabrv,References}

\onecolumn
\appendices

\section{Hyperparameters used in Experiments}
\renewcommand{\arraystretch}{1.2}
\begin{table}[h!]
\centering
\captionsetup{justification=centering}
\caption{Hyperparameters of PPO Agents}
\begin{tabular}{cccc}
&  \multicolumn{3}{c}{ \textbf{Agents}}\\ 
\cline{2-4} \\
\textbf{Hyperparameters}    & PPO & PPO+CTS & PPO+ICM \\
\hline
Policy                  &  CNN              &   CNN             &   CNNLstm         \\
Timesteps               &   1e8             &     1e8           &      2e8          \\
Horizon                 &   256             &     256           &      64           \\
Num. Minibatch          &   8               &      8            &      8            \\
GAE $(\lambda)$         &   0.95            &     0.95          &     0.99          \\
Discount $(\gamma)$     &   0.99            &     0.99          &     0.999         \\
Learning Rate $(\alpha)$& $5\times 10^{-4}$ & $5\times 10^{-4}$ & $5\times 10^{-4}$ \\
Num. Epochs             &   3               &       3           &       4           \\
Entropy Coeff.          &   0.01            &     0.01          &     0.001         \\
VF Coeff.               &   0.5             &     0.5           &     0.5           \\
Clipping Param.         &   0.2             &     0.2           &     0.1           \\
Max Grad. Norm.         &   0.5             &     0.5           &     0.5           \\
Num. of Actors          &   16              &     16            &     32            \\
CTS Beta $(\beta)$      &    -              &       0.05        &      -            \\
CTS Filter              &    -              &    L-shaped       &      -            \\
ICM State Features      &    -              &       -           &     256           \\
ICM Beta $(\beta)$      &    -              &       -           &     0.2           \\
\end{tabular}
\label{table:e:agents}
\end{table}

\renewcommand{\arraystretch}{1.2}
\begin{table}[h!]
\centering
\captionsetup{justification=centering}
\caption{Hyperparameters of APF Techniques}
\begin{tabular}{ccc}
\textbf{Hyperparameters}    & APFCTS & APFICM \\
\hline
$pos_{cap}$             &   ~0.4              &     ~0.1          \\
$neg_{cap}$             &   -0.4              &    -0.4           \\
APF Beta $(\beta)$      &   ~0.01             &     ~0.01         \\
\end{tabular}
\label{table:e:apf}
\end{table}

\end{document}